\pdfoutput=1
\documentclass[11pt]{article}

\usepackage[final]{acl}

\usepackage{times}
\usepackage{latexsym}
\usepackage{amsmath}
\usepackage{amssymb}
\usepackage{txfonts}
\usepackage{graphicx}
\usepackage{booktabs}
\usepackage{subcaption}

\usepackage[T1]{fontenc}
\usepackage[utf8]{inputenc}

\usepackage{microtype}

\usepackage{inconsolata}

\newcommand{\condprob}[2]{\mathsf{P}( #1 \mid #2 )}

\newcommand{\magenta}[1] {\textcolor{magenta}{#1}}

\title{Leading Whitespaces of Language Models' Subword Vocabulary Pose a Confound for Calculating Word Probabilities}

\author{Byung-Doh Oh \\
  Center for Data Science \\
  New York University \\
  \texttt{oh.b@nyu.edu} \\\And
  William Schuler \\
  Department of Linguistics \\
  The Ohio State University \\
  \texttt{schuler.77@osu.edu} \\}

\begin{document}
\maketitle
\begin{abstract}
Predictions of word-by-word conditional probabilities from Transformer-based language models are often evaluated to model the incremental processing difficulty of human readers.
In this paper, we argue that there is a confound posed by the most common method of aggregating subword probabilities of such language models into word probabilities.
This is due to the fact that tokens in the subword vocabulary of most language models have leading whitespaces and therefore do not naturally define stop probabilities of words.
We first prove that this can result in distributions over word probabilities that sum to more than one, thereby violating the axiom that $\mathsf{P}(\Omega) = 1$.
This property results in a misallocation of word-by-word surprisal, where the unacceptability of the end of the current word is incorrectly carried over to the next word.
Additionally, this implicit prediction of word boundaries incorrectly models psycholinguistic experiments where human subjects directly observe upcoming word boundaries.
We present a simple decoding technique to reaccount the probability of the trailing whitespace into that of the current word, which resolves this confound.
Experiments show that this correction reveals lower estimates of garden-path effects in transitive/intransitive sentences and poorer fits to naturalistic reading times.
\end{abstract}

\section{Introduction}
Language models (LMs), which are trained to make predictions about upcoming words, are at the core of many natural language processing (NLP) applications.
While most contemporary applications involve generating text by sampling from the LMs' conditional probability distribution, the magnitudes of the probabilities they assign to each word in a given sentence have been important from two perspectives.
The first is from the perspective of LM interpretability, which aims to study their predictions and the linguistic knowledge encoded in their representations.
A well-established paradigm in this line of research is what has been dubbed ``targeted syntactic evaluation'' \citep{linzenetal16, gulordavaetal18, marvinlinzen18}, in which probabilities of critical words in minimal pairs (e.g.~grammatical vs.~ungrammatical sentences) are compared. 

Moreover, in cognitive modeling, conditional probabilities from LMs are used to model the word-by-word reading times of human subjects, often under the theoretical link that the contextual predictability of a word determines its processing difficulty \citep{hale01, levy08}.
Recent work in this line of research has evaluated surprisal estimates (i.e.~negative log probabilities) from LMs and has shown that surprisal from larger Transformer-based model variants are less predictive of naturalistic reading times \citep{ohschuler23tacl, shainetal24, steueretal23} and that surprisal greatly underpredicts the processing difficulty of garden-path constructions \citep{vanschijndellinzen21, arehallietal22, huangetal24}.

As such, while the use of word-by-word probabilities from LMs is popular in computational linguistics research, we argue that there is a confound for calculating them correctly that has gone unaddressed.
This confound is posed by subword tokenization schemes \citep[e.g.~byte-pair encoding;][]{sennrichetal15} that are used to define the token-level vocabulary for training most contemporary LMs \citep[e.g.][]{llama24, gemini24, jiangetal23}.
For languages that use whitespace orthography, these subword tokenization schemes often build the whitespace character directly into the front of the tokens, thereby resulting in \textit{leading} whitespaces.
As a consequence, the stop probability of a word (i.e.~the probability of the \textit{trailing} whitespace) is never explicitly calculated, and therefore the sum over the probabilities of all possible whitespace words can exceed one.

We propose a simple and efficient decoding method that reaccounts the probability of the trailing whitespace into that of the current word, which resolves this confound.
Regression results show that this correction reveals significantly lower surprisal-based estimates of garden-path effects in transitive/intransitive sentences and poorer fits of LM surprisal to naturalistic reading times.

\section{Confound From Leading Whitespaces and Whitespace-Trailing Decoding}
This section provides a proof that the leading whitespaces of the LMs' subword vocabulary result in inconsistent word probabilities, describes a related confound, and proposes a simple decoding method for addressing it.

\subsection{Proof of Inconsistent Word Probabilities}
On languages that use whitespace orthography, the vocabulary $V$ defined by the subword tokenization scheme consists of the set of tokens that begin with a whitespace $V_B$, and the set of tokens that do not begin with a whitespace $V_I$.\footnote{$V = V_B \cup V_I$, and $V_B \cap V_I = \varnothing$. The subscripts respectively represent the `beginning' and `inside' of a whitespace-delimited word.}
In the context of next-word prediction, the sample space of a whitespace-delimited word is $\Omega = \{x_{1..n} \mid x_1 {\in} V_B, \, x_{2..n} {\in} V_I, \, n {\in} \mathbb{N}\}$, where $n$ is the total number of subword tokens in each whitespace word as determined by the tokenizer.

\paragraph{Theorem 1}
Leading whitespaces of the LMs' subword vocabulary can result in word probabilities that violate the \citet{kolmogorov33} axiom that $\mathsf{P}(\Omega) = 1$.

\paragraph{Existence Proof}
Let $\mathsf{P}(x_1{=}j_1) = 1$ and $\condprob{x_2{=}j_2}{x_1{=}j_1} = 1$, where $j_1 {\in} V_B$ and $j_2 {\in} V_I$.
It follows from the chain rule of conditional probabilities that:
\begin{align}
    \mathsf{P}(x_1{=}j_1, x_2{=}j_2) &= \mathsf{P}(x_1{=}j_1) \cdot \condprob{x_2{=}j_2}{x_1{=}j_1} \notag \\ &= 1 \cdot 1 = 1.
\end{align}
If word probabilities are simply defined as the product of the probabilities of the tokens within those words, then $\mathsf{P}(x_1{=}j_1) + \mathsf{P}(x_1{=}j_1, x_2{=}j_2) > 1$, $\mathsf{P}(\Omega) > 1$, and therefore the axiom is violated. \hfill $\blacksquare$

For example, given the minimal pair \textit{I was a matron in France} and \textit{I was a mat in France}, where \textit{matron} is more likely than \textit{mat}, the LM tokenizes the two sentences as follows and calculates the conditional probability of each token.\footnote{In the context of the LM's tokens, `\textvisiblespace ' is used to denote the explicit whitespace character that is part of the token, and whitespace is used to delimit subword tokens.}
\begin{align}
    & \textit{I \ \textvisiblespace was \ \textvisiblespace a \ \textvisiblespace mat \ ron \ \textvisiblespace in \ \textvisiblespace France} \label{ex:sameword} \\
    & \textit{I \ \textvisiblespace was \  \textvisiblespace a \ \textvisiblespace mat \ \textvisiblespace in \ \textvisiblespace France} \label{ex:newword}
\end{align}

\begin{figure}[t!]
    \centering
    \begin{subfigure}[t]{0.48\textwidth}
        \includegraphics[width=\textwidth]{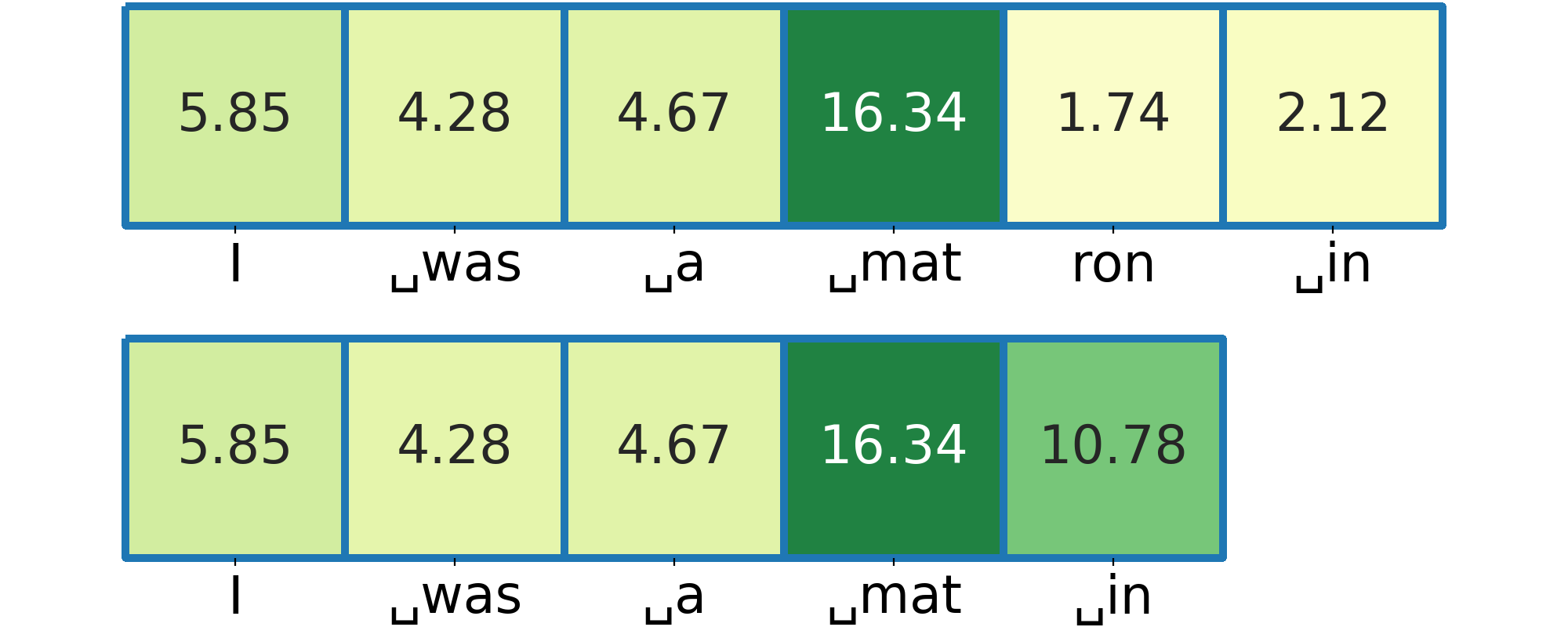}
        \caption{Surprisal values calculated with leading whitespaces.}
        \label{fig:overview_wl}
    \end{subfigure} \par \bigskip
    
    \begin{subfigure}[t]{0.48\textwidth}
        \includegraphics[width=\textwidth]{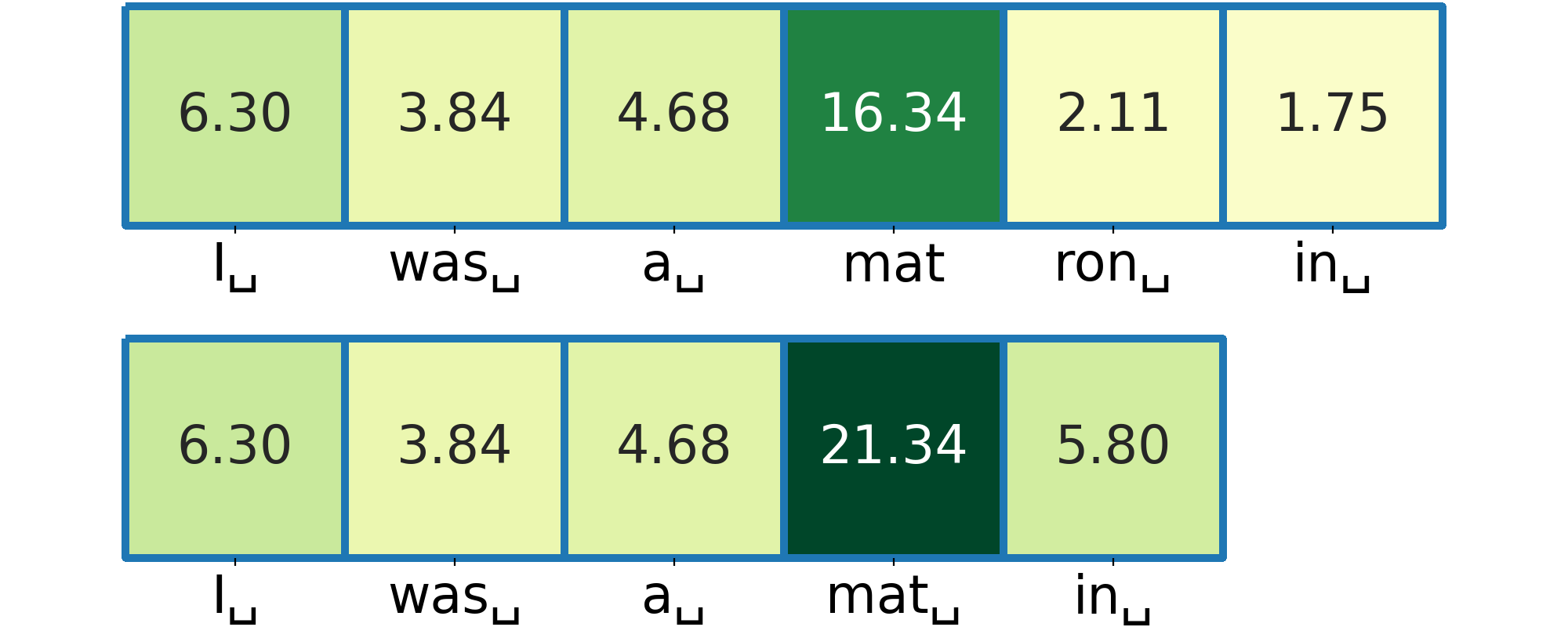}
        \caption{Surprisal values calculated with trailing whitespaces.}
        \label{fig:overview_wt}
    \end{subfigure}
    \caption{Surprisal values calculated for the partial sentences \textit{I was a matron in} and \textit{I was a mat in} using the GPT-2 XL LM \citep{radfordetal19}, with leading whitespaces (top; standard practice) and trailing whitespaces (bottom; proposed in this work).}
    \label{fig:overview}
\end{figure}

The presence of leading whitespaces results in an incorrect allocation of word-by-word surprisal.
As can be seen in Example \ref{ex:sameword}, due to this tokenization, $\condprob{\textit{\textvisiblespace mat ron}}{\textit{I \textvisiblespace was \textvisiblespace a}}$ is factorized into $\condprob{\textit{\textvisiblespace mat}}{\textit{I \textvisiblespace was \textvisiblespace a}} \cdot \condprob{\textit{ron}}{\textit{I \textvisiblespace was \textvisiblespace a \textvisiblespace mat}}$, and therefore it follows that $\condprob{\textit{\textvisiblespace mat ron}}{\textit{I \textvisiblespace was \textvisiblespace a}} \leq \condprob{\textit{\textvisiblespace mat}}{\textit{I \textvisiblespace was \textvisiblespace a}}$, despite the fact that \textit{matron} is more acceptable than \textit{mat} in the above context.
Instead, part of the `unacceptability' of \textit{mat} is incorrectly carried over to $\condprob{\textit{\textvisiblespace in}}{\textit{I \textvisiblespace was \textvisiblespace a \textvisiblespace mat}}$, where \textit{\textvisiblespace in} competes for probability mass against the highly likely \textit{ron} (Figure \ref{fig:overview_wl}).

\subsection{Confound: Incompatibility With Psycholinguistic Experimental Paradigms}
Additionally, the presence of leading whitespaces in subword tokens makes the LM's predictions incompatible with the self-paced reading paradigm, in which human readers directly observe the upcoming word boundary.
\begin{align}
    \textcolor{lightgray}{\textit{The cat sat on the}} \textit{ mat} \textcolor{lightgray}{\textit{ ...}}\label{ex:sprnewword}
\end{align}
In Example \ref{ex:sprnewword}, when human readers see \textit{mat}, they know that the next keystroke will reveal a new whitespace-delimited word (analogous to observing that the next token will be in $V_B$) and not transform it into e.g.~\textit{matron} (analogous to observing that the next token will be in $V_I$).
In contrast, LMs define a probability distribution over both $V_B$ and $V_I$ after the token \textit{mat} in the sequence \textit{The cat sat on the mat}.

While this confound is more apparent in the self-paced reading paradigm, this is also a potential confound for studying data collected through the typical eye-tracking paradigm.
This is because native speakers of languages with whitespace orthographies have been shown to be sensitive to the location of upcoming whitespaces through parafoveal processing and utilize this information to plan eye movements \citep{pollatsekrayner82, rayneretal98, pereaacha09}.
Therefore, although information about word boundaries is not directly built into the design of the paradigm, it can be argued that human subjects engaged in the eye-tracking paradigm also face little uncertainty about upcoming word boundaries.

\subsection{Proposed Solution: Whitespace-Trailing Decoding}
This inconsistency and confound can be resolved by reaccounting the probability of the \textit{trailing} whitespace as part of the word's probability, in lieu of that of the \textit{leading} whitespace as LMs currently do (Examples \ref{ex:sameword} and \ref{ex:newword}).
To this end, we propose whitespace-trailing (WT) decoding.
Given a word $w_{t+1}$ that consists of subword tokens $x_{n_{t}+1..n_{t+1}}$, where $n_{t}$ is the total number of subword tokens in the word sequence $w_{1..t}$, and $x_{n_{t}+1} {\in} V_B$, and $x_{n_{t}+2..n_{t+1}} {\in} V_I$, WT decoding reallocates the probability of the leading whitespace of each word to its previous word:\footnote{See Appendix \ref{sec:wt_proof} for the proof that WT decoding results in consistent word probabilities. However, we note that WT decoding does not resolve other issues with subword units that may be addressed by re-training LMs with different tokenization schemes \citep[e.g.][]{nairresnik23}, which can nonetheless be expensive. Concurrent work by \citet{pimentelmeister24} points out this same issue and also proposes WT decoding.}
\begin{multline}
    \mathsf{P}( w'_{t+1} \mid w'_{1..t}) = \label{eq:wtprob} \\ \condprob{w_{t+1}}{w_{1..t}} \cdot \frac{\condprob{x_{n_{t+1}+1} {\in} V_B}{w_{1..t+1}}}{\condprob{x_{n_{t}+1} {\in} V_B}{w_{1..t}}}.
\end{multline}
For instance, applying Equation \ref{eq:wtprob} to Example \ref{ex:newword} yields:
\begin{multline}
    \condprob{\textit{mat\textvisiblespace}}{\textit{I\textvisiblespace\,was\textvisiblespace\,a\textvisiblespace}} = \label{eq:wtexample} \\
    \condprob{\textit{\textvisiblespace mat}}{\textit{I\,\textvisiblespace was\,\textvisiblespace a}} \cdot \frac{\condprob{\textit{\textvisiblespace}}{\textit{I\,\textvisiblespace was\,\textvisiblespace a\,\textvisiblespace mat}}}{\condprob{\textit{\textvisiblespace}}{\textit{I\,\textvisiblespace was\,\textvisiblespace a}}}.
\end{multline}

As WT decoding simply involves the factorization of whitespace probabilities by marginalizing over tokens in $V_B$ and rearranging them, it requires no modifications to the LM and minimal overhead.
Additionally, the joint probability of the entire sequence, and therefore metrics like perplexity, changes minimally by a factor of the probability of the final trailing whitespace with WT decoding.

As can be seen in Figure \ref{fig:overview_wt}, incorporating the probabilities of trailing whitespaces correctly differentiates between \textit{matron} and \textit{mat} in this context, and removes the inherent relationship between the two probabilities that holds with leading whitespaces.
Additionally, the `unacceptability' of \textit{mat} that was incorrectly carried over to \textit{\textvisiblespace in} in Example \ref{ex:newword} is now reflected in $\condprob{\textit{mat\textvisiblespace}}{\textit{I\textvisiblespace \ was\textvisiblespace \ a\textvisiblespace}}$.

LM probabilities with trailing whitespaces are also better aligned with the self-paced reading paradigm where the upcoming word boundaries are directly observed.
For example, the calculation of $\condprob{\textit{mat\textvisiblespace}}{\textit{The\textvisiblespace \ cat\textvisiblespace \ sat\textvisiblespace \ on\textvisiblespace \ the\textvisiblespace}}$ precludes the prediction of tokens in $V_I$ directly after \textit{mat}, which correctly reflects the fact that the next keystroke in Example \ref{ex:sprnewword} will reveal a new whitespace word.

\section{Experiment 1: Surprisal-Based Estimates of Garden-Path Effects} \label{sec:exp1}
Equation \ref{eq:wtexample} shows that WT decoding will result in an increase (or decrease) in probability to the extent that the next token is likely to be in $V_B$ proportional to the extent that the first token of the current word was likely to be in $V_B$.
The first experiment demonstrates that the confound posed by leading whitespaces affects surprisal-based estimates of garden-path effects in transitive/intransitive sentences \citep{mitchell87, gorrell91}, which is caused by syntactic disambiguation that takes place at the critical word (highlighted in magenta).
\begin{align}
    & \textit{\small After the doctor left the room \magenta{turned} very dark ...} \label{ex:amb_npz}
\end{align}
The same critical word in the control counterpart is thought to be easier to process, as the verb \textit{left} is disambiguated by the comma.
\begin{align}
    \displaystyle
    & \textit{\small After the doctor left, the room \magenta{turned} very dark ...} \label{ex:unamb_npz}
\end{align}

\subsection{Procedures}
We estimated surprisal-based garden-path effects from GPT-2 model variants \citep{radfordetal19} with and without WT decoding, using the data and following the procedures of \citet{huangetal24}.
First, to estimate a linking function between LM surprisal and human reading times, linear mixed-effects regression (LMER) models with the following formula were fit to self-paced reading times ($n=995,814$) of filler items from the Provo Corpus \citep{lukechristianson18} for each variant:

\begin{small}
\begin{verbatim}
RT ~ surp + surp_prev1 + surp_prev2 + s(length) + 
     freq + freq_prev1 + freq_prev2 + s(index) +
     (1 | subject) + (1 | item),
\end{verbatim}
\end{small}
where \texttt{length} is word length in characters, \texttt{index} is the position of the word within the sentence, and the \texttt{frequency} predictors were log-transformed.

These modeling choices assume a linear relationship between surprisal and reading times \citep{shainetal24, wilcoxetal23}, and that surprisal and log frequency from two previous words have a lingering influence on the current word (spillover effects).
These LMER models were subsequently used to predict word-by-word reading times (in ms) for 24 items in the ambiguous condition (Example \ref{ex:amb_npz}) and the unambiguous control condition (Example \ref{ex:unamb_npz}) of the transitive/intransitive construction, which were read by 2,000 subjects ($n{=}15,915$).

The increase in the predicted reading times of the disambiguating critical word and two subsequent words due to the increase in surprisal across conditions was estimated using LMER models with the following formula to quantify the magnitude of surprisal-based garden-path effects at each word:\footnote{Both the `filler item' and `reading time increase' LMER models have been simplified from the specifications in \citet{huangetal24} due to convergence issues.}

\begin{small}
\begin{verbatim}
pred_RT ~ condition + (1 | subject) + (1 | item).
\end{verbatim}
\end{small}

\begin{figure}[t!]
\centering
    \includegraphics[width=0.48\textwidth]{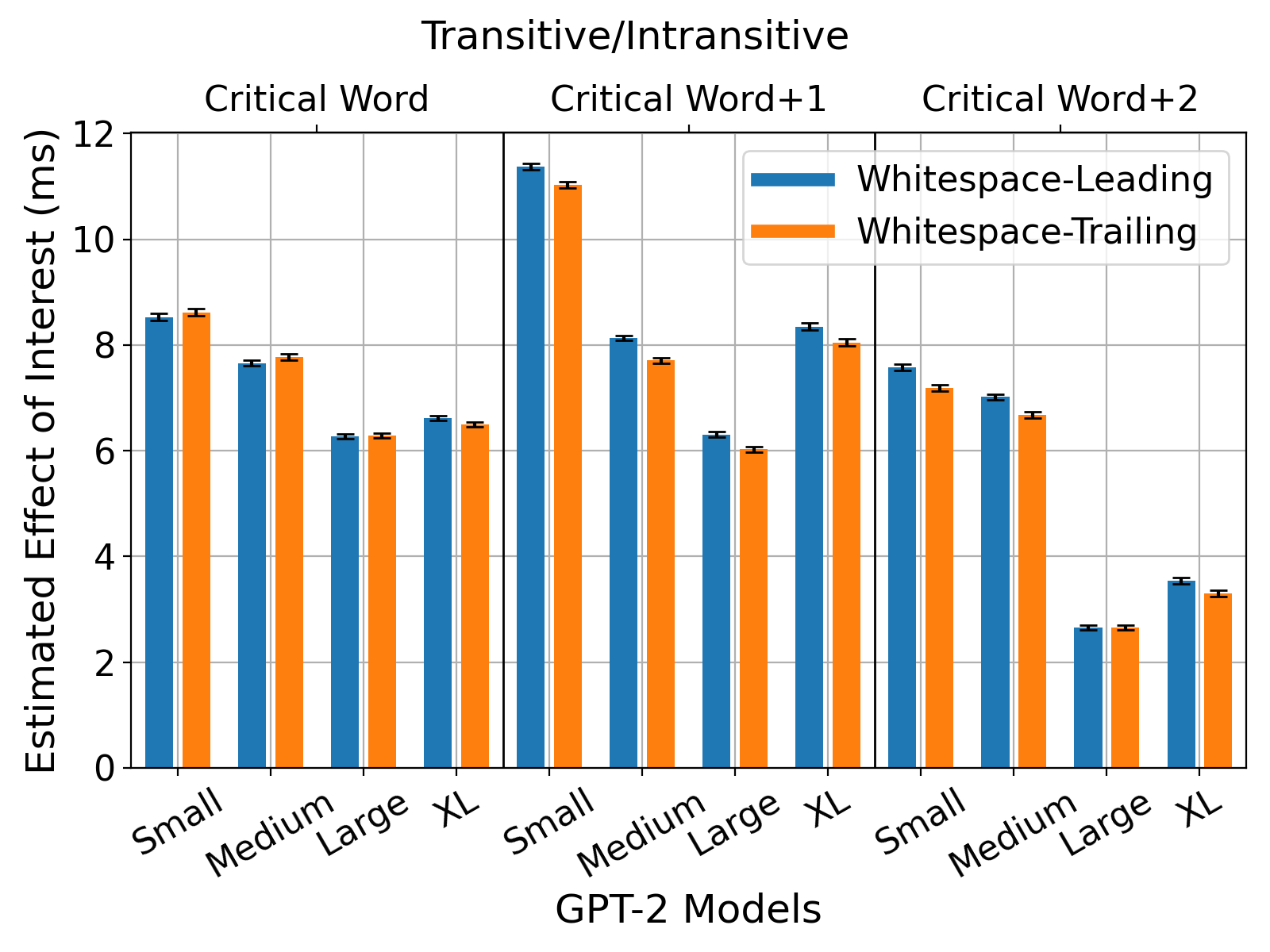}
\caption{Estimated effects of interest at each region for the transitive/intransitive garden-path construction, using GPT-2 surprisal with and without WT decoding. Error bars represent 95\% confidence intervals.}
\label{fig:gp_results}
\end{figure}

\begin{figure*}[ht!]
    \includegraphics[width=0.4975\textwidth]{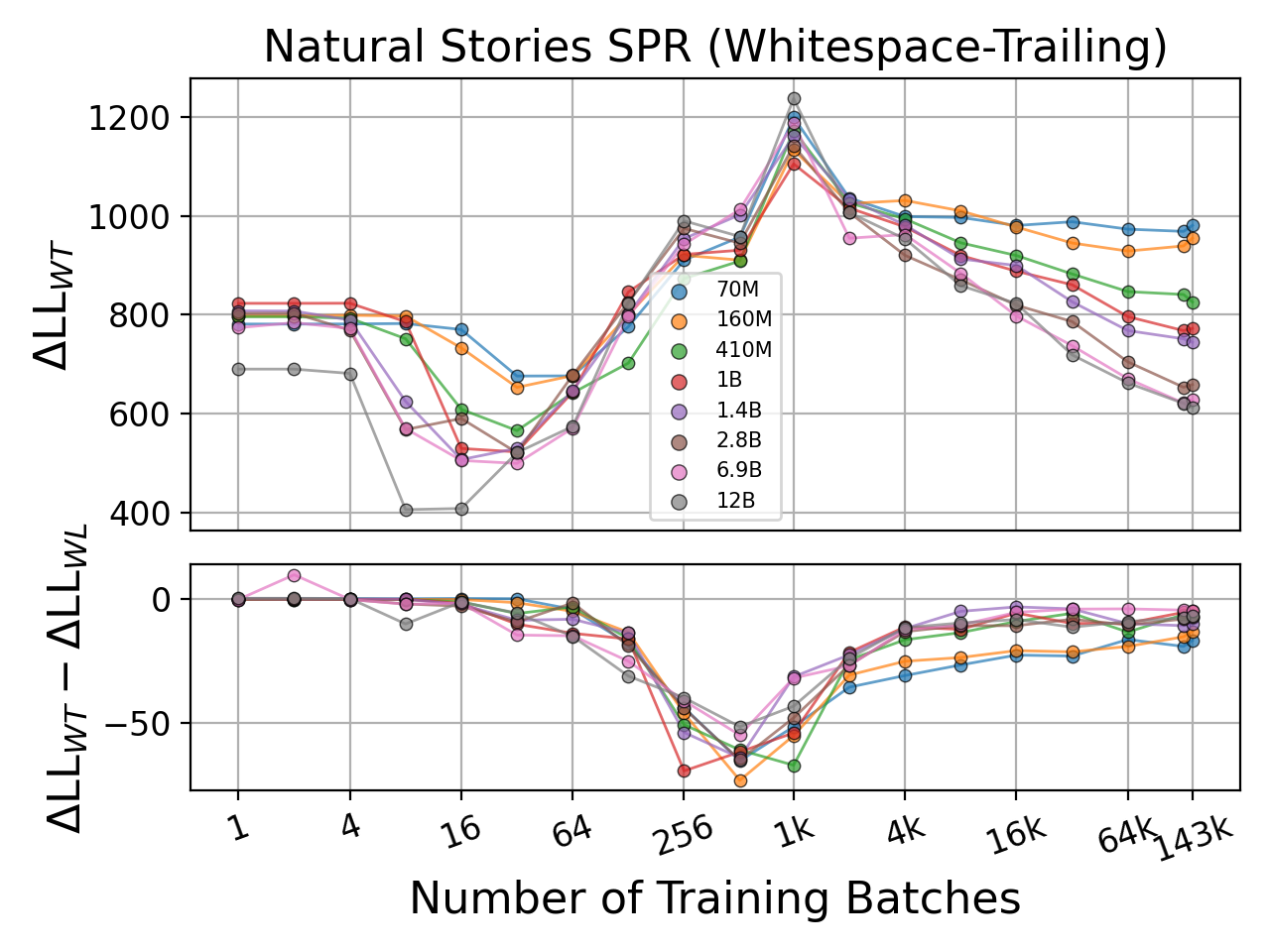}
    \includegraphics[width=0.4975\textwidth]{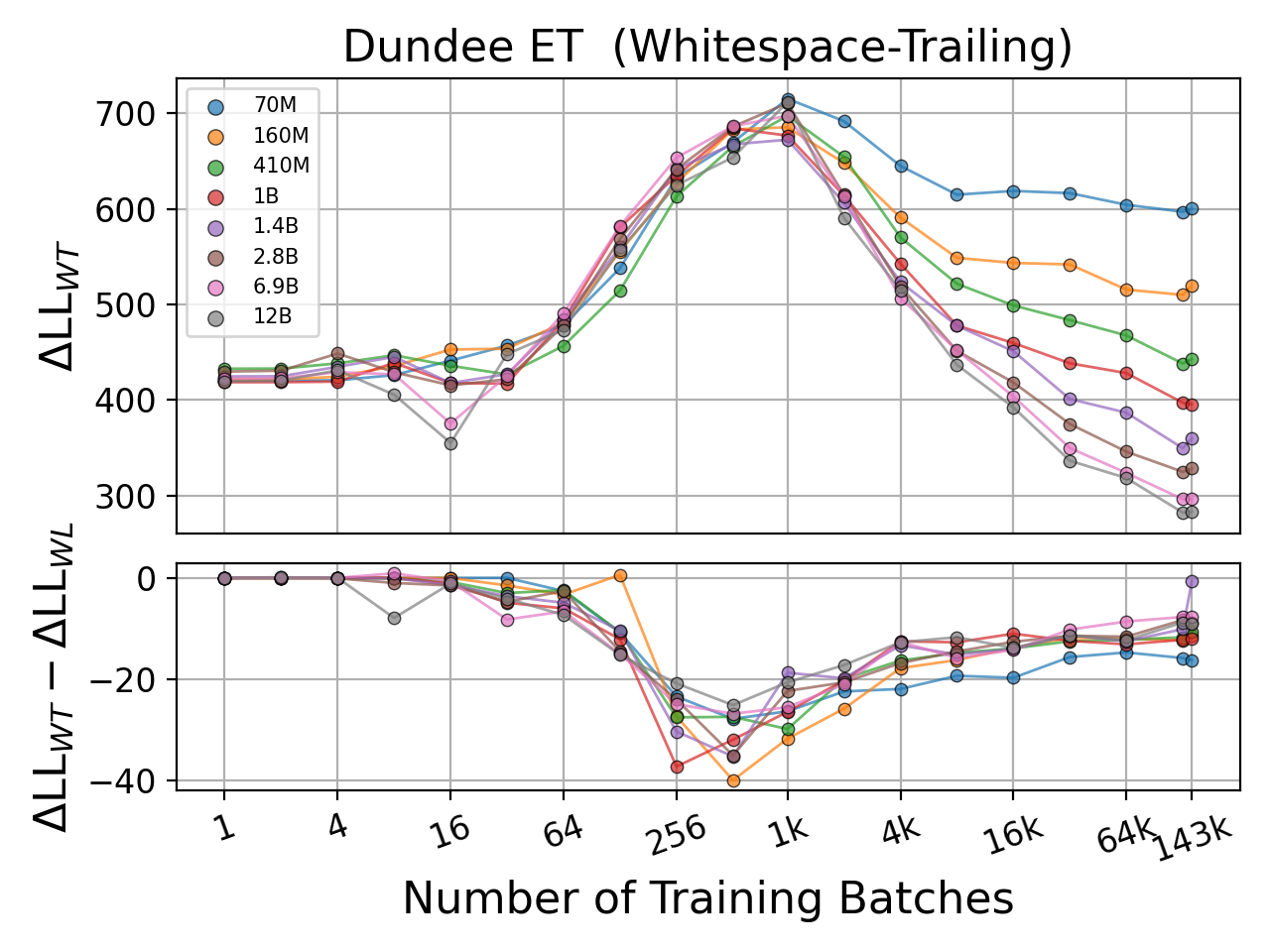}
\caption{Increase in regression model log-likelihood due to including surprisal estimates from Pythia LM variants calculated with WT decoding (top) and the resulting change in regression model log-likelihood (bottom). See Appendix \ref{sec:nrt_results_wl} for results from surprisal estimates calculated without WT decoding.}
\label{fig:nrt_results_wt}
\end{figure*}

\subsection{Results}
The results in Figure \ref{fig:gp_results} show addressing the confound posed by subword tokenization through WT decoding lowers the estimated magnitude of garden-path effects in the first and second spillover regions, the difference in which is significant at $p<0.05$ level for all comparisons except GPT-2 Large in the second spillover region.
This is due to the decrease in surprisal at the critical region of the ambiguous condition (\textit{turned}), as the probability of its unlikely preceding whitespace\footnote{The LMs strongly expect a comma right after \textit{room}.} is reaccounted by the previous word (\textit{room}).
The resulting decrease in surprisal difference across conditions at the critical region is carried over to the two spillover regions.
At the critical region itself, however, this decrease is not observed as the increase in surprisal difference of its previous word (\textit{room}) cancels it out.
Such lower estimates suggest that the underestimation of human-like garden-path effects is more severe than previously reported.

\section{Experiment 2: Fit of Surprisal to Naturalistic Reading Times} \label{sec:exp2}
The second experiment evaluates how addressing the confound posed by leading whitespaces affects the fit of LM surprisal to naturalistic reading times.

\subsection{Procedures}
The experimental procedures closely follow those of \citet{ohschuler23emnlp}, who evaluated surprisal estimates from Pythia LM variants \citep{bidermanetal23} with different model sizes and training data amounts on self-paced reading (SPR) times from the Natural Stories Corpus \citep{futrelletal21} and go-past durations (GPD) from the Dundee Corpus \citep{kennedyetal03}.\footnote{See Appendix \ref{sec:preprocessing} for the data preprocessing procedures.}
LMER models with the following formulae were respectively fit to the Natural Stories and Dundee corpora, whose likelihoods were then subtracted from those of the baseline LMER models without surprisal to calculate the increase in log-likelihood due to surprisal, or $\Delta$LL:

\begin{small}
\begin{verbatim}
log(SPR) ~ surp + length + index +
           (surp + length + index + 1 | subject) +
          (1 | subject:sentid)
log(GPD) ~ surp + length + index + slength + pfix +
           (surp + length + index + slength + pfix +
          1 | subject) + (1 | sentid),
\end{verbatim}
\end{small}
where \texttt{slength} is the saccade length, \texttt{pfix} is whether the previous word was fixated, and \texttt{sentid} is the index of the sentence within each corpus.
These procedures were repeated with and without WT decoding to calculate $\Delta$LL$_{WT}$ and $\Delta$LL$_{WL}$ respectively, and the change in fit to reading times as a result of addressing the confound was calculated.

\subsection{Results}
Figure \ref{fig:nrt_results_wt} shows that surprisal estimates calculated with WT decoding results in poorer fits to naturalistic reading times, especially for LMs that have seen around 256 to 1,000 batches of training data on both corpora.
Nonetheless, the peak in $\Delta$LL at around 1,000 training batches\footnote{The change in log-likelihood ($\Delta$LL$_{WT} - \Delta$LL$_{WL}$) at 1,000 training batches is significant at $p<0.001$ level on both corpora by a permutation test of aggregated squared errors.} and the adverse effect of model size at the end of LM training \citep{ohschuler23emnlp} are replicated.
In contrast to these results, \citet{pimentelmeister24} report small improvements on the same two corpora as a result of applying WT decoding to fully trained Pythia LMs.
We conjecture this is due to different regression modeling procedures involving different baseline predictors.

\section{Conclusion}
This work calls attention to an inconsistency and a confound that is inherent in word probabilities calculated from LMs trained with subword tokenization.
These are posed by the fact that tokens have leading whitespaces in most models, meaning that the stop probability of a whitespace word is never explicitly calculated, which can result in word probability distributions whose sum exceeds one.
We proposed WT decoding as a solution for these issues, and demonstrated that addressing them reveals lower surprisal-based estimates of transitive/intransitive garden-path effects and poorer fits of LM surprisal to naturalistic reading times.
Other targeted syntactic constructions and naturalistic reading time corpora may similarly show systematic changes to word probabilities.

More generally, addressing these issues will have the biggest impact on probabilities of words neighboring low-probability whitespaces, such as those at potential phrasal/clausal boundaries where LMs will likely predict a punctuation mark.
These issues will also be more pronounced for LMs that are not able to predict word boundaries accurately, such as those trained on smaller amounts of data.
Therefore, future studies using LM word probabilities for interpretability and cognitive modeling research should control for them through WT decoding.\footnote{Code for implementing WT decoding is available at: \url{https://github.com/byungdoh/wt_decoding}.}

\section*{Acknowledgments}
We thank the ARR reviewers and the area chair for their helpful comments.
This work was supported by the National Science Foundation (NSF) grant \#1816891.
All views expressed are those of the authors and do not necessarily reflect the views of the NSF.
Computations for this work were partly run using the \citet{osc87}.

\section*{Limitations}
The confound in the connection between word-by-word conditional probabilities of Transformer-based language models and human reading times identified in this work is supported by experiments using language model variants trained on English text and data from human subjects that are native speakers of English.
Therefore, the confound identified in this work may not generalize to other languages, in particular those that do not use whitespace orthography.
Additionally, this work is concerned with the use of language models as cognitive models of human sentence processing, and therefore does not relate to their use in natural language processing applications, such as text generation, summarization, or question answering.

\section*{Ethics Statement}
This work used data collected as part of previously published research \citep{huangetal24, lukechristianson18, futrelletal21, kennedyetal03}.
Readers are referred to the respective publications for more information on the data collection and validation procedures.
As this work focuses on studying the connection between conditional probabilities of language models and human sentence processing, its potential negative impacts on society appear to be minimal.

\bibliography{custom}
\newpage
\appendix

\onecolumn
\section{Proof of Consistent Word Probabilities With Whitespace-Trailing Decoding} \label{sec:wt_proof}

\paragraph{Theorem 2}
Applying whitespace-trailing decoding results in word probabilities that satisfy the \citet{kolmogorov33} axiom that $\mathsf{P}(\Omega) = 1$.

\paragraph{Proof}
In the context of predicting $w_{t+1}$ given $w_{1..t}$, the sample space is $\Omega = \{x_{n_{t}+1..n_{t+1}} \mid x_{n_{t}+1} {\in} V_B, \, x_{n_{t}+2..n_{t+1}} {\in} V_I, \, \{n_{t}, n_{t+1}\} {\subset} \mathbb{N}, \, n_{t+1} {>} n_t\}$, where $n_{t}$ is the total number of subword tokens in the word sequence $w_{1..t}$, and $n_{t+1}$ is the total number of subword tokens in the word sequence $w_{1..t+1}$.
Therefore, $\mathsf{P}(\Omega)$ is the total sum of word probabilities when $n_{t+1}-n_{t}=1, 2, 3, ...$ .

The sum of word probabilities according to Equation \ref{eq:wtprob} when $n_{t+1}-n_{t} = 1$ is:
\begin{multline}
    \sum_{j_{1} {\in} V_B} \condprob{x_{n_{t}+1} {=} j_{1}}{w_{1..t}} \cdot \frac{\condprob{x_{n_{t}+2} {\in} V_B}{x_{n_{t}+1} {=} j_{1}, w_{1..t}}}{\condprob{x_{n_{t}+1} {\in} V_B}{w_{1..t}}} = \frac{\condprob{x_{n_{t}+1} {\in} V_B, x_{n_{t}+2} {\in} V_B}{w_{1..t}}}{{\condprob{x_{n_{t}+1} {\in} V_B}{w_{1..t}}}} \\ = \condprob{x_{n_{t}+2} {\in} V_B}{x_{n_{t}+1} {\in} V_B, w_{1..t}}.
\end{multline}
More generally, the sum of word probabilities when $n_{t+1}-n_{t} \geq 2$ is:
\begin{multline}
    \sum_{\substack{j_{1} {\in} V_B \\ j_{2..(n_{t+1}-n_{t})} {\in} V_I}} \!\! \condprob{x_{{n_{t}+1}..{n_{t+1}}}{=}j_{1..(n_{t+1}-n_{t})}}{w_{1..t}} \cdot \frac{\condprob{x_{n_{t+1}+1} {\in} V_B}{x_{{n_{t}+1}..{n_{t+1}}}{=}j_{1..(n_{t+1}-n_{t})}, w_{1..t}}}{\condprob{x_{n_{t}+1} {\in} V_B}{w_{1..t}}} \\ = \condprob{x_{n_{t}+2..n_{t+1}} {\in} V_I, x_{n_{t+1}+1} {\in} V_B}{x_{n_{t}+1} {\in} V_B, w_{1..t}}.
\end{multline}
$\mathsf{P}(\Omega)$ can then be calculated as the following series that sums over disjoint subspaces of $\Omega$:
\begin{align}
    \mathsf{P}(\Omega) = \ & \condprob{x_{n_{t}+2} {\in} V_B}{x_{n_{t}+1} {\in} V_B, w_{1..t}} \ + \notag \\ & \condprob{x_{n_{t}+2} {\in} V_I, x_{n_{t}+3} {\in} V_B}{x_{n_{t}+1} {\in} V_B, w_{1..t}} \ + \notag \\ & \condprob{x_{n_{t}+2} {\in} V_I, x_{n_{t}+3} {\in} V_I, x_{n_{t}+4} {\in} V_B}{x_{n_{t}+1} {\in} V_B, w_{1..t}} \ + \notag \\ & \condprob{x_{n_{t}+2} {\in} V_I, x_{n_{t}+3} {\in} V_I, x_{n_{t}+4} {\in} V_I, x_{n_{t}+5} {\in} V_B}{x_{n_{t}+1} {\in} V_B, w_{1..t}} \ + \notag \\ & ... \ ,
\end{align}
which approaches $\condprob{x_{n_{t}+2} {\in} V_B}{x_{n_{t}+1} {\in} V_B, w_{1..t}} + \condprob{x_{n_{t}+2} {\in} V_I}{x_{n_{t}+1} {\in} V_B, w_{1..t}} = 1$ in the limit. \hfill $\blacksquare$

\section{Preprocessing Procedures for Naturalistic Reading Time Corpora}
\label{sec:preprocessing}

The Natural Stories Corpus \citep{futrelletal21} provides self-paced reading times from 181 subjects that read 10 English stories (10,256 words), which were filtered to exclude those shorter than 100 ms or longer than 3000 ms, those of sentence-initial and -final words, and those from subjects who answered fewer than four comprehension questions correctly.
Approximately 50\% of the observations (384,905 observations) selected based on the sum of the subject index and the sentence index was used to fit the LMER models and calculate $\Delta$LL.

The Dundee Corpus \citep{kennedyetal03} provides fixation durations from 10 subjects that read 67 English newspaper editorials (51,501 words), which were filtered to exclude those from unfixated words, those of words following saccades longer than four words, and those of sentence/document/line/screen-initial and -final words.
Again, approximately 50\% of the observations (98,115 observations) selected based on the sum of the subject index and the sentence index was used to fit the LMER models and calculate $\Delta$LL.

\section{Increase in Regression Model Log-Likelihood Without WT Decoding}
\label{sec:nrt_results_wl}

\begin{figure*}[ht!]
    \includegraphics[width=0.4975\textwidth]{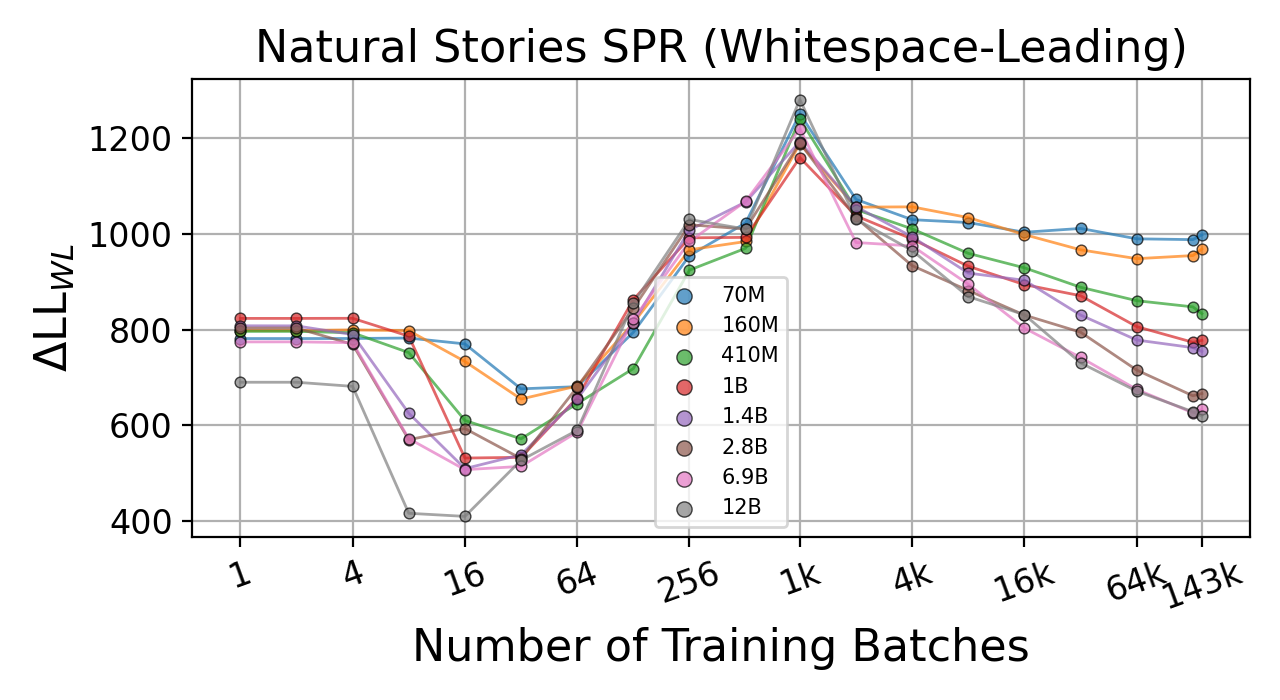}
    \includegraphics[width=0.4975\textwidth]{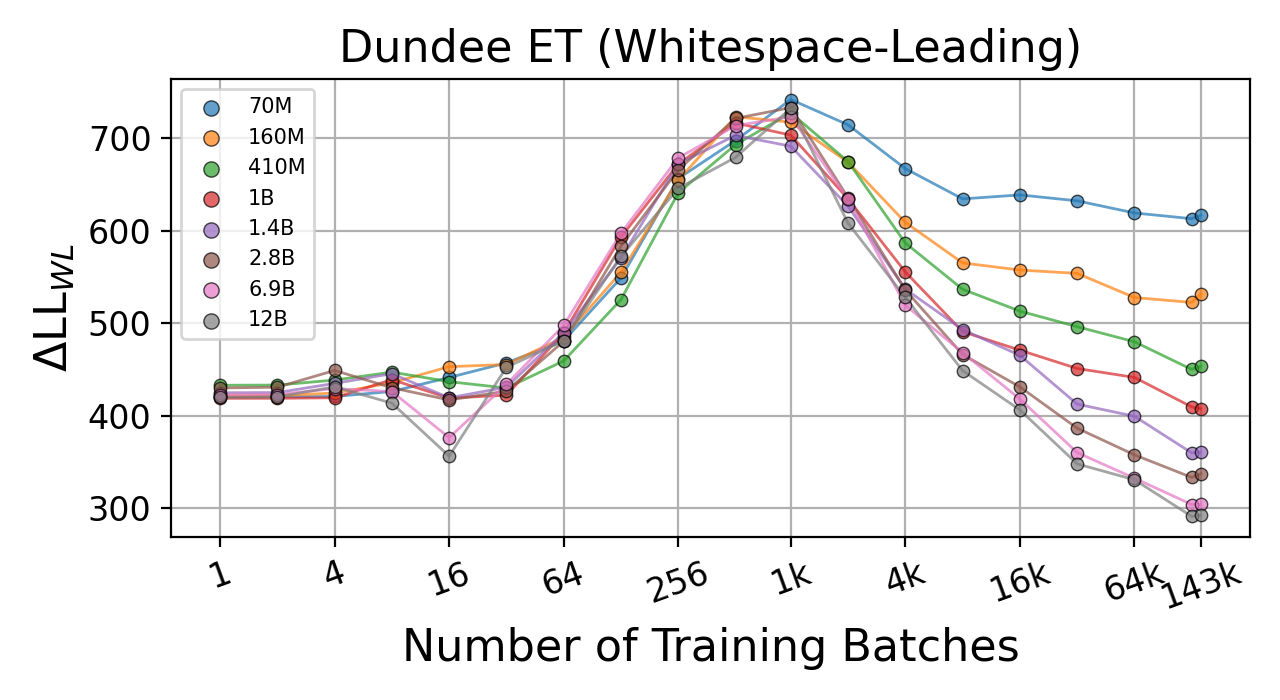}
\caption{Increase in regression model log-likelihood due to including surprisal estimates from Pythia LM variants calculated without WT decoding.}
\label{fig:nrt_results_wl}
\end{figure*}

The increase in regression model log-likelihood due to including surprisal estimates from Pythia LM variants calculated without WT decoding can be found in Figure \ref{fig:nrt_results_wl}.

\end{document}